\documentclass[conference]{IEEEtran}
\IEEEoverridecommandlockouts
\usepackage{cite}
\usepackage{amsmath,amssymb,amsfonts}
\usepackage{algorithmic,wrapfig}
\usepackage{graphicx}
\usepackage{textcomp}
\usepackage{xcolor}
\usepackage{url}

\newcommand{\Yashesh}[1]{\textcolor{blue}{Yashesh: #1}}

\newcommand{\seethis}[1]{\textcolor{red}{#1}}

\usepackage{caption}
\usepackage{subcaption}
\usepackage{comment}
\usepackage{relsize}
\usepackage{multirow}
\usepackage{array}
\newcolumntype{H}{>{\setbox0=\hbox\bgroup}c<{\egroup}@{}}

\def\BibTeX{{\rm B\kern-.05em{\sc i\kern-.025em b}\kern-.08em
    T\kern-.1667em\lower.7ex\hbox{E}\kern-.125emX}}
\begin{document}

\title{Evaluating Nonlinear Decision Trees for Binary Classification Tasks with Other Existing Methods
}

\author{\IEEEauthorblockN{Yashesh Dhebar\IEEEauthorrefmark{1}, Sparsh Gupta\IEEEauthorrefmark{2}, and Kalyanmoy Deb\IEEEauthorrefmark{1}}
\IEEEauthorblockA{\IEEEauthorrefmark{1}Computational Optimization and Innovation (COIN) Laboratory, Michigan State University, East Lansing, Michigan, USA \\
\{dhebarya, kdeb\}@egr.msu.edu}
\IEEEauthorblockA{\IEEEauthorrefmark{2}Department of Mechanical Engineering,
Indian Institute of Technology, Kanpur, Uttar Pradesh, India,
sparshg@iitk.ac.in}
}


\maketitle

\begin{abstract}
Classification of datasets into two or more distinct classes is an important machine learning task. Many methods are able to classify binary classification tasks with a very high accuracy on test data, but cannot provide any easily interpretable explanation for users to have a deeper understanding of reasons for the split of data into two classes. In this paper, we highlight and evaluate a recently proposed nonlinear decision tree approach with a number of commonly used classification methods on a number of datasets involving a few to a large number of features. The study reveals key issues such as effect of classification on the method's parameter values, complexity of the classifier versus achieved accuracy, and interpretability of resulting classifiers.   
\end{abstract}


\section{Introduction}
The task of a binary classification algorithm is to arrive at a classifier involving one or more features from a set of two-labelled dataset, so that the resulting classifier is able to correctly classify unseen test datasets of similar type into two classes with near 100\% accuracy. The classifier can be a mathematical function of features, or a network in which features act as input to the network and a binary output reveals the class of a data point, or a decision tree in which a data point flows from root node to internal nodes according to the decisions made at each node and ending up with a class identification at one of the leaf nodes. Each representation (a mathematical function, a network or a decision tree) can be {\em simple\/}, involving fewer terms and structure, or complex. However, it is well understood that the complexity of a classifier and its achievable testing accuracy are closely linked. A classifier which is simple most likely cannot be very accurate and vice versa. Fortunately, most classification methods are involved with one or more algorithmic parameters that can be tuned to achieve a desired above-mentioned accuracy-complexity trade-off. 

An important matter which is getting a lot of attention in the classification literature is the interpretability of obtained classifiers. Besides accurately classifying new data into its true class, the users are getting more interested in learning how the classifier is able to classify a data into its true class with an easy-to-explain logic. If a classifier has a complex structure (to achieve a high enough classification accuracy), the resulting classifier may be too complex to interpret and explain. Hence, a classification method capable of producing a good balance between accuracy and interpretability is desired.

In this paper, we consider a number of popular classification methods -- a linear decision tree (CART), support vector machines (SVMs), generalized additive models (GAMs), genetic programming (GP), and a recently proposed nonlinear decision tree (NLDT) approach. We discuss their working principles in brief and provide their advantages and disadvantages in Section~\ref{sec:principles}. After providing the effect of their parameters on the obtained accuracy-complexity trade-off, we compare them on 19 different binary classification problems (described in Section~\ref{sec:problems}) having two to 500 features in Section~\ref{sec:results}. Finally, conclusions are drawn in Section~\ref{sec:conclusions}. 

\section{Existing Binary Classification Methods}
\label{sec:principles}
In this section, we provide a brief description of a few popular existing classification methods pertaining to binary classification tasks. 

\subsection{Classification and Regression Trees (CART)}
\begin{figure}[hbt]
    \centering
    \includegraphics[width = 0.9\linewidth]{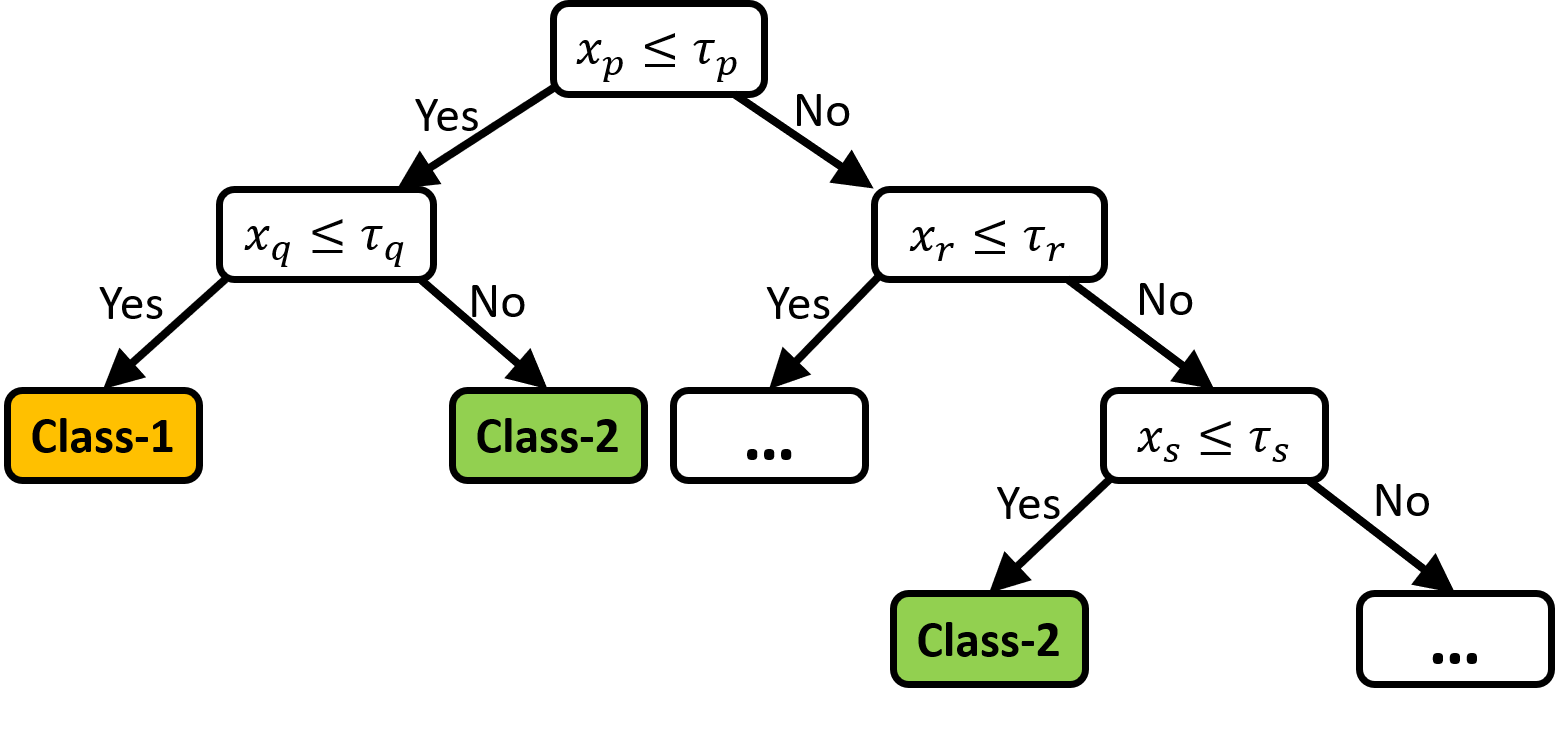}
    \caption{A CART decision tree splitting the flow of a data into one of the two branches, finally leading to a class identification at its terminal leaf nodes.}
    \label{fig:cart_illustration}
\end{figure}

Classification and regression trees or CART have been thought of as a popular choice, since the resulting classifier assumes the structure of a \emph{decision tree}. Decision trees make decision using a logical hierarchical representation, which is also common to the way in which a human mind operates. The overall structure is represented in an inverted tree format, with the root node at the top and leaf nodes as the terminals. The data in the root node undergoes recursive binary splitting \cite{quinlan1986induction, quinlan2014c4} to create child nodes in the decision tree. One restriction of the CART approach is that splits in decision trees are axis parallel in nature and operate on only one feature (i.e. $x_i \le \tau_i^*$), as shown in Figure~\ref{fig:cart_illustration}.

The spilt rule $x_i \le \tau$ splits the data in the conditional node ($P$) (the node where split is occurring) into two non-overlapping subsets: left child node ($L$) and right child node ($R$). The quality of split is computed by using an {\em impurity\/} metric, like the Gini score, entropy, or others. An impurity metric quantifies the purity (or impurity) of data distribution in a given node:
\begin{equation}
    {\rm Gini} = 1 - \sum_i^c \frac{N_i}{N},
    \label{eq:gini}
\end{equation}
where $c$ is the number of classes (which is two in our case), $N$ is the total number of data points in the node and $N_i$ is the number of data points in the given node belonging to class $i$. The quality of split $(S)$ can then be computed using the following equation:
\begin{equation}
    S = \frac{N_L}{N_P}{\rm Gini}(L) + \frac{N_R}{N_P}{\rm Gini}(R),
    \label{eq:split_quality}
\end{equation}
where $N_P$ is the total number of points in the given parent node undergoing a split, $N_L$ and $N_R$ are number of points belonging to left child node (for which $x_i \le \tau$ is TRUE) and right child node (for which $x_i \le \tau$ is FALSE), respectively. The optimal feature $x_i$ and its optimal threshold value $\tau_i$ are determined using a greedy algorithm, or through a univariate optimization method. The $(x_i, \tau_i)$ combination giving the lowest $S$-value (Eq.~\ref{eq:split_quality}) is chosen to conduct the split. A recursive algorithm ID3 \cite{quinlan1986induction} or C4.5 \cite{quinlan2014c4,breiman2017classification} is employed to grow the tree.

The tree is allowed to grow up to a prespecified maximum depth when the node under consideration meets one of the termination criteria. The nodes that do not undergo any further split are referred to as \emph{leaf} nodes. The leaf node is assigned with a \emph{class} based on the distribution of data within the node. Since the split rule at each conditional node assumes a very simple linear structure, i.e. $x_i \le \tau$, many  splits are required for a complex classification task, thereby resulting into a complicated decision tree topology, which may not be fathomable by a human. 

Some advantages and disadvantages of the CART method for binary  classification are listed below:

\noindent {\bf Advantages:}
\begin{itemize}
    \item Fast to train.
    \item Easily interpretable rules (linear and each rule involves only one of the features) in each node.
    \item Many source codes and packages available for quick implementation.
\end{itemize}

\noindent {\bf Disadvantages:}
\begin{itemize}
    \item The execution requires a number of tunable parameters: (i) maximum depth of the tree, (ii) total number of splits, (iii) threshold impurity level and (iv) minimum number of classified data points in a node for terminating any further split and declaring it as a leaf node. Available codes come with default values, which may not produce a desired accuracy or end up with a huge decision tree. 
    \item The method has a tendency to overfit the training data, leading to poor performance on test data. Pruning and other methods, like bagging and boosting, are suggested \cite{kearns1999boosting, zhang2010vertical, ke2017lightgbm} to overcome this effect.
    \item The tree eventually grows as a result of many hierarchical successive spitting and becomes topologically very complex for humans to fathom.
    \item Clearly, the method is not suitable for datasets which require a complex, nonlinear, and linked feature relationships for achieving an accurate classification.  
\end{itemize}

In our experiments in this paper, we use Matlab's \emph{fitctree()} routine with its default parameter settings to generate CART based classifiers. 

\newcommand{\boldx}{\mbox{${\mathbf x}$}}
\newcommand{\boldw}{\mbox{${\mathbf w}$}}
\subsection{Support Vector Machines (SVMs)}
\begin{figure}[hbt]
    \centering
    \includegraphics[width = 0.65\linewidth]{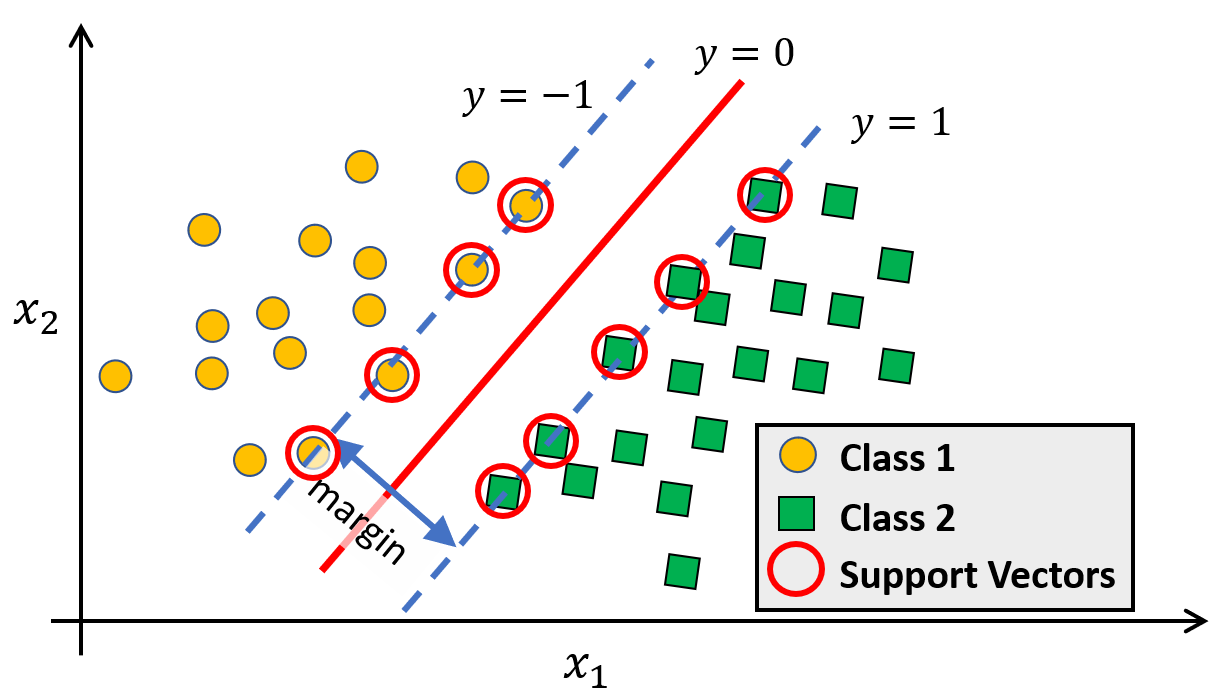}
    \caption{SVM on separable datasets with a hard margin.}
    \label{fig:svm_hard_margin}
\end{figure}

\begin{figure}[hbt]
    \centering
    \includegraphics[width = 0.65\linewidth]{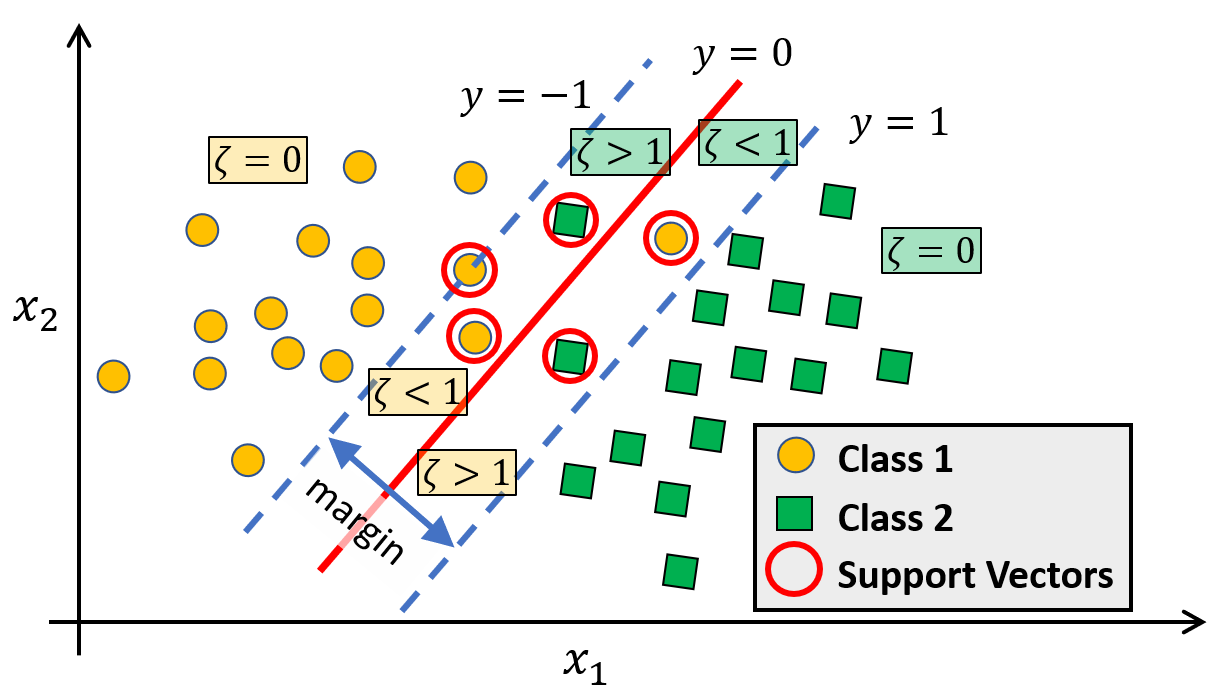}
    \caption{SVM with non-separable datasets with a soft margin.}
    \label{fig:svm_soft_margin}
    \vspace{-10pt}
\end{figure}

For a separable dataset, support vector machine (SVM) algorithm attempts to derive a decision boundary in the form of a single mathematical equation as shown below:
\begin{equation}
    y(\mathbf{x}) = \mathbf{w}^T\boldsymbol{\phi}(\mathbf{x}) + b,
    \label{eq:svm_main}
\end{equation}
where $\boldsymbol{\phi}(\mathbf{x})$ is a set of feature transformation functions which can be either linear or non-linear functions of feature vector $\boldx$, $\boldw$ is a weight vector and $b$ is a bias term. 
A conceptual understanding of SVM is provided in Figure~\ref{fig:svm_hard_margin}. 
For a binary classification task involving class labels $t = -1$ or $t = 1$, an optimal hyper-surface is derived by maximizing the margin between two classes, as shown by $y=0$ line in the figure. Points with $y\leq -1$ belong to one class and points with $y\geq 1$ belong to another class. The points which fall on $y=1$ and $y=-1$ are called support vectors, as they alone decide the classifier.   However, for non-separable datasets, such as the scenario shown in Figure~\ref{fig:svm_soft_margin}, a \emph{soft} margin approach is used to allow some data points within $|y| < 1$ (margin) while training the SVM. These points are also declared as support vectors in addition to the points on the margin.

To identify the classifier and the support vectors, the underlying optimization problem is solved:
\begin{equation}
    \begin{array}{rl}
     \text{Minimize:}    & \frac{1}{2}||w||^2 + C\mathlarger{\sum}_{i = 1}^N\zeta_i,\\
     \mbox{subject to}: & t_i(\mathbf{w}^T\boldsymbol{\phi}(\mathbf{x}_i) + b) \ge 1 - \zeta_i, \\
     & \zeta_i \ge 0, \quad i = 1,2,\dots,N,
    \end{array}
    \label{eq:svm_optim}
\end{equation}
where $t_i$ is the \emph{true} class label (either 1 or -1) of the datapoint, $\zeta_i$ is the distance of $i$-th data point from its representative margin, thus $\zeta_i = \text{max}\left[0, 1 - t_iy(\mathbf{x}_i)\right]$ (where value of $y(\mathbf{x}_i)$ is estimated from Eq.~\ref{eq:svm_main}). 
$C$ is a penalty parameter which is used to enhance \emph{generalizability} by compromising with \emph{training accuracy}. It is also aimed to balance the complexity of the classifier (described with the number of non-zero terms of $\boldw$) and soft support vectors within the margin and is an important parameter. With lower values of $C$, broader margin (with some misclassification of training datapoints) is achieved while for large values of $C$, misclassification of training datapoints is heavily penalized and so narrower margin is achieved.



Using a kernel trick \cite{bishop2006pattern}
$k(\mathbf{x}_p, \mathbf{x}_q) = \boldsymbol{\phi}(\mathbf{x}_p)^T\boldsymbol{\phi}(\mathbf{x}_q)$ Eq.~\ref{eq:svm_main} is transformed into the following:
\begin{equation}
    y(\mathbf{x}) = \sum_{i = 1}^{N}a_it_ik(\mathbf{x}, \mathbf{x}_i) + b,
    \label{eq:svm_kernel}
\end{equation}
where $a_i$ is a Lagrange multiplier which is obtained by converting the optimization problem of maximizing the margin (Eq.~\ref{eq:svm_optim}) to a dual Lagrangian representation \cite{bishop2006pattern}:
\begin{equation}
    \begin{array}{rl}
         \text{Min:} & L(\mathbf{a}) = \mathlarger{\sum}_{i = 1}^Na_i - \frac{1}{2}\mathlarger{\sum}_{i = 1}^N\mathlarger{\sum}_{j = 1}^Na_ia_jt_it_jk(\mathbf{x}_i,\mathbf{x}_j), \\
         \text{s.t.} & a_i \in [0,C],\quad  \mathlarger{\sum}_{i=1}^Na_it_i = 0.
    \end{array}
    \label{eq:svm_dual}
\end{equation}
Classical gradient based algorithms can then be employed to find $a_i$. In Eq.~\ref{eq:svm_kernel}, data points $\mathbf{x}_i$ for which $a_i = 0$  do not contribute in the equation of the split rule (Eq.~\ref{eq:svm_kernel}) and data points for which $a_i > 0$ are called support vectors for the SVM classifier and they dictate the overall length of the classifier's equation (Eq.~\ref{eq:svm_kernel}). The penalty parameter $C$ has to be tuned to efficiently derive the decision boundary. Lower value of $C$ makes the classifier more generalizable. $C = \infty$ (hard margin) attempts to achieve near $100\%$ training accuracy and hence is prone to overfitting. In our case, we use scikit-learn's \cite{scikit-learn} SVM module and set $C = 1,000$. We use RBF (or Gaussian) kernel function. 
Table~\ref{tab:svm_results} shows results for various settings of $C$ on some datasets considered in our study. 

\noindent {\bf Advantages:}
\begin{itemize}
    \item Good in many classification tasks and scales well with dimension of the dataset.
    \item Since a classical optimization solver is employed to solve the Lagrangian dual problem (Eq.~\ref{eq:svm_dual}), the training is fast.
    \item Can generalize well through an appropriate choice of $C$.
    \item Many source codes and packages are available for rapid implementation of SVM on various languages like python \cite{scikit-learn} and Matlab.
\end{itemize}

\noindent {\bf Disadvantages:}
\begin{itemize}
    \item The penalty parameter $C$ acts as a regularization parameter and needs to be properly identified and tuned while working on different datasets.
    \item The knowledge regarding \emph{separability} of datapoints belonging to different classes is required to properly tune $C$ and in practical problems, this information is not available.
    \item The kernel function $k(\mathbf{x}_p, \mathbf{x}_q)$ (Eq.~\ref{eq:svm_kernel}) needs to be chosen.
    
    \item Since only one rule is found through SVM, the resulting rule might involve many terms, thereby making the overall classifier uninterpretable.
\end{itemize}


\begin{table*}[hbt]
    \centering
    \caption{SVM Result for different values of penalty parameter $C$. For each dataset, the first row represents the testing accuracy and the second row represents complexity (number of support vectors). $C = 1000$ gives overall best performance.}
    \label{tab:svm_results}
    
    \begin{tabular}{|c|c|c|c|c|c|c|}\hline
    \textbf{Pen. Param.} & \textbf{DS1} & \textbf{DS2} & \textbf{DS3} & \textbf{DS4} & \textbf{Truss} & \textbf{WeldedBeam}\\\hline
    ${C = 1}$ & 
    
    \begin{tabular}{c}
         $ 94.75 \pm 1.97 $ \\ $ 191.94 \pm 4.38 $ 
    \end{tabular} & 
    \begin{tabular}{c}
        $ 95.24 \pm 0.00 $ \\ $ 16.56 \pm 0.80 $ 
    \end{tabular} & \begin{tabular}{c}
        $ 96.93 \pm 1.87 $ \\ $ 64.68 \pm 2.16 $
    \end{tabular} & 
    \begin{tabular}{c}
       $ 45.20 \pm 4.24 $ \\ $ 138.76 \pm 1.99 $
    \end{tabular} & \begin{tabular}{c}
        $ 77.31 \pm 2.16 $ \\ $ 343.76 \pm 9.51 $
    \end{tabular} & \begin{tabular}{c}
         $ 98.83 \pm 0.70 $ \\ $ 47.26 \pm 3.00 $
    \end{tabular}\\\hline
    ${C = 10}$ & 
    \begin{tabular}{c}
        $ 98.42 \pm 1.16 $ \\ $ 58.70 \pm 2.90 $ 
    \end{tabular} & \begin{tabular}{c}
         $ 95.24 \pm 0.00 $ \\ $ 30.46 \pm 0.64 $
    \end{tabular} & \begin{tabular}{c}
         $ 99.32 \pm 0.70 $ \\ $ 26.68 \pm 1.88 $ 
    \end{tabular} & 
    \begin{tabular}{c}
        $ 68.77 \pm 4.00 $ \\ $ 262.60 \pm 4.76 $ 
    \end{tabular} & \begin{tabular}{c}
         $ 81.29 \pm 2.19 $ \\ $ 258.52 \pm 10.85 $
    \end{tabular} & \begin{tabular}{c}
         $ 99.53 \pm 0.42 $ \\ $ 17.86 \pm 1.73 $ 
    \end{tabular}\\\hline
    ${C = 1,000}$ & 
    \begin{tabular}{c}
        $\mathbf{ 99.88 \pm 0.33 }$ \\ $ \mathbf{ 8.36 \pm 0.87 }$
    \end{tabular} & \begin{tabular}{c}
         $\mathbf{ 99.70 \pm 0.50 }$ \\ $\mathbf{ 8.56 \pm 0.67 }$
    \end{tabular} & \begin{tabular}{c}
         $\mathbf{ 99.75 \pm 0.58 }$ \\ $\mathbf{ 10.60 \pm 0.92 }$ 
    \end{tabular} & 
    \begin{tabular}{c}
        $\mathbf{ 96.63 \pm 1.35 }$ \\ $\mathbf{ 56.70 \pm 3.13 }$
    \end{tabular} & \begin{tabular}{c}
         $\mathbf{ 88.54 \pm 1.60 }$ \\ $\mathbf{ 176.22 \pm 7.87 }$ 
    \end{tabular} & \begin{tabular}{c}
         $\mathbf{ 99.63 \pm 0.38 }$ \\ $\mathbf{ 7.88 \pm 0.86 }$ 
    \end{tabular}\\\hline
    \end{tabular}%
    \vspace{5pt}
    \begin{tabular}{|c|c|c|c|c|cH|}\hline
    \textbf{Pen. Param.} & \textbf{m-DS1} & \textbf{m-DS2} & \textbf{m-DS3} & \textbf{Cancer-10} & \textbf{Cancer-30} & \textbf{{\color{red}{m-ZDT-500}}}\\\hline
    ${C = 1}$ & 
    \begin{tabular}{c}
    $ 99.77 \pm 0.67 $ \\ $ 70.22 \pm 2.23 $
    \end{tabular} & \begin{tabular}{c}
       $ 95.24 \pm 0.00 $ \\ $ 16.18 \pm 0.59 $
    \end{tabular} & \begin{tabular}{c}
       $ 99.97 \pm 0.23 $ \\ $ 36.54 \pm 1.72 $
    \end{tabular}  & \begin{tabular}{c}
        $\mathbf{ 97.15 \pm 1.08 }$ \\ $ 69.98 \pm 6.51 $
    \end{tabular} & \begin{tabular}{c}
       $ 90.83 \pm 1.83 $ \\ $ 106.88 \pm 4.44 $ 
    \end{tabular} & 
    \begin{tabular}{c}
        \textbf{--}  \\ comp
    \end{tabular}\\\hline
    ${C = 10}$ & 
    \begin{tabular}{c}
        $\mathbf{ 100.00 \pm 0.00 }$ \\ $ 26.42 \pm 1.46 $
    \end{tabular} & \begin{tabular}{c}
         $ 98.89 \pm 0.85 $ \\ $ 14.40 \pm 0.89 $ 
    \end{tabular} & \begin{tabular}{c}
         $\mathbf{ 100.00 \pm 0.00 }$ \\ $ 12.60 \pm 0.98 $ 
    \end{tabular} & \begin{tabular}{c}
         $ 95.98 \pm 1.13 $ \\ $ 56.22 \pm 6.43 $ 
    \end{tabular} & \begin{tabular}{c}
         $ 91.94 \pm 1.36 $ \\ $ 81.66 \pm 4.54 $ 
    \end{tabular}  & 
    \begin{tabular}{c}
        \textbf{--} \\ comp 
    \end{tabular}\\\hline
    ${C = 1,000}$ & 
    \begin{tabular}{c}
        $ 99.93 \pm 0.33 $ \\ $\mathbf{ 7.38 \pm 0.75 }$
    \end{tabular} & \begin{tabular}{c}
         $\mathbf{ 99.97 \pm 0.22 }$ \\ $\mathbf{ 5.34 \pm 0.55 }$
    \end{tabular} & \begin{tabular}{c}
         $\mathbf{ 100.00 \pm 0.00 }$ \\ $\mathbf{ 8.82 \pm 1.01 }$
    \end{tabular} & \begin{tabular}{c}
         $ 95.23 \pm 1.09 $ \\ $\mathbf{ 52.36 \pm 4.91 }$
    \end{tabular} & \begin{tabular}{c}
        $\mathbf{ 95.08 \pm 1.65 }$ \\ $\mathbf{ 58.74 \pm 5.18 }$
    \end{tabular} & 
    \begin{tabular}{c}
        \textbf{--}  \\ comp 
    \end{tabular}\\\hline
    \end{tabular}
\end{table*}

\subsection{Generalized Additive Models (GAMs)}
For a binary classification task involving two classes: Class 1 ($y = 0$) and Class 2 ($y = 1$), the GAM based classifier \cite{hastie1990generalized, wood2017generalized} estimates the probability of a data point belonging to class $y = 1$ (i.e. $P(y = 1|\mathbf{x})$)\footnote{probability of datapoint belonging to other class (i.e. $y = 0$) will be $1-\hat{y}$.} as $\hat{y}(\mathbf{x})$ using the following equation
\begin{equation}
    \hat{y}(\mathbf{x}) = \frac{1}{1 + \mathlarger{e^{-g(\mathbf{x})}}},
    \label{eq:gam_sigmoid}
\end{equation}
where $g(\mathbf{x})$ is referred to as \emph{link function} \cite{larsen2015gam}. The link function $g(\mathbf{x})$ in GAM is expressed as a sum of non-linear functions as shown below:
\begin{equation}
    g(\mathbf{x}) = f_1(\mathbf{x}) + f_2(\mathbf{x}) + \dots + f_M(\mathbf{x}) + \beta_0,
    \label{eq:gams}
\end{equation}
where $\beta_0$ is a constant and $f_i(\mathbf{x})$ are scalar valued nonlinear functions. The functional form of $f_i(\mathbf{x})$ and total number of such nonlinear functions is pre-specified by the user. Modelling of link function $g(\mathbf{x})$ using Eq.~\ref{eq:gams} makes GAMs more generalizable than its precursor: generalized linear models (GLMs) \cite{nelder1972generalized}, which involves only linear terms.

 In our experiments, we use penalized B-splines to model non-linearity of each feature separately (i.e. referring to Eq.~\ref{eq:gams}, $f_i(\mathbf{x}) = s_i(x_i)$). Thus, the $g$-function in our case is given by
 \begin{equation}
 \begin{array}{rl}
      & g(\mathbf{x}) = s_1(x_1) + s_2(x_2) \dots s_d(x_d) + \beta_0,\\
     \textrm{where:} & s_i(x_i) = \mathlarger{\sum_{j=1}}^{K_i}B_{j}^{(q_i)}(x_i)\beta_j = \mathbf{B}'_i(x_i)\boldsymbol{\beta}_i.
\end{array}
\label{eq:gam_splines}
 \end{equation}
 Here, $s_i(x_i)$ denotes a spline function corresponding to $i$-th feature, $B_{j}^{(q_i)}(x_i)$ indicates the basis function of order $q_i$, $\beta_j$ are scalar coefficients and $K_i$ is the total number of basis functions used to model the spline. The order of spline (i.e. $q_i$) and the number of basis-functions $K_i$ is user-specified. 
 
Once the structure of link function $g(\mathbf{x})$ is specified, an optimization algorithm is invoked to learn parameters corresponding to basis functions $\beta_j^{(q_i)}(x_i)$ and  coefficients $\beta_j$  with an objective to minimize the error between the estimated value of probability ($\hat{y}(\mathbf{x})$ Eq.~\ref{eq:gam_sigmoid}) and the actual $y$ values across the dataset. To make the resulting model more generalize and simple, a second-order smoothing is employed. Thus, using Eq.~\ref{eq:gam_sigmoid} and \ref{eq:gam_splines}, the overall optimization problem translates to minimizing the following function:
\vspace{-5pt}
{\small\begin{equation}
    \vspace{-3pt}
    \textrm{Min:}\ F(\mathbf{B'},\boldsymbol{\beta})\!=\! \mathlarger{\sum}_{i = 1}^N\!(y_i - \hat{y}_i(\mathbf{B'}\boldsymbol{\beta}))^2 
    + \mathlarger{\sum}_{j = 1}^d\lambda_j\!\mathlarger{\int}\!\!(s''_j(x_j|_{\mathbf{B}'_j\boldsymbol{\beta}_j}))^2dx_j,
    \label{eq:gam_optimization}
\end{equation}}
where $y_i$ is the actual class of the $i$-th datapoint (which can have value of either 0 or 1) and $\hat{y}_i$ is the probability of $i$-th point belonging to class $y = 1$ $\left( \text{i.e. } P(y = 1|\mathbf{x_i})\right)$ as predicted by the GAM classifier using Eq.~\ref{eq:gam_sigmoid}. $\lambda_j$ are the penalty parameters which are prespecified. In our case, we use $\lambda_j = 0.6$ for all features. The rule complexity of a GAM classifier can be tuned using $\lambda_j$, where higher values of $\lambda_j$ imposes heavy penalty on non-linearities with more than second order. Additionally, the complexity can also be controlled by regulating the \emph{degree} ($q_i$) and \emph{number of basis-functions} $K_i$ (Eq.~\ref{eq:gam_splines}). In our experimental setup, we conduct series of experiments using different combinations of $(K_i,q_i)$ to model splines for each feature. Values of $K$ and $q$  are picked from the one listed in Table~\ref{tab:gam_parameters}.

\begin{figure}[hbtp]
\begin{subfigure}[b]{0.5\linewidth}
\hspace*{-2mm}\includegraphics[width = 1.1\linewidth]{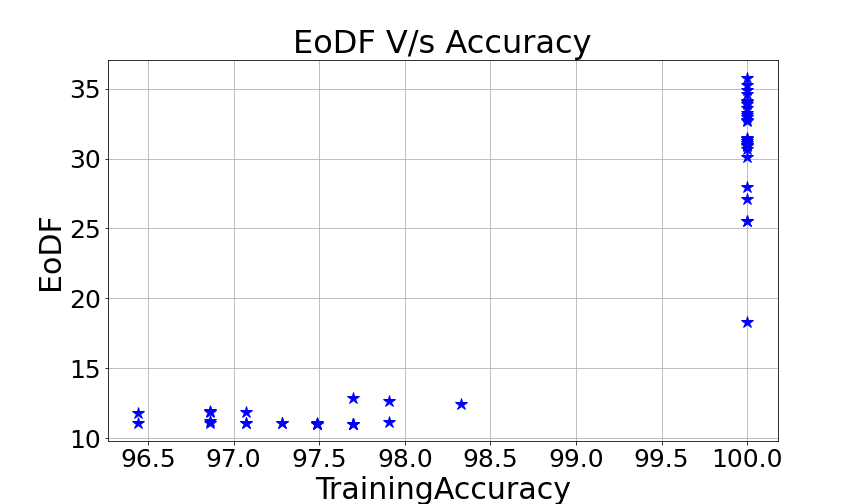}
\caption{Training Accuracy}
\end{subfigure}%
~
\begin{subfigure}[b]{0.5\linewidth}
\hspace*{-3mm}\includegraphics[width = 1.1\linewidth]{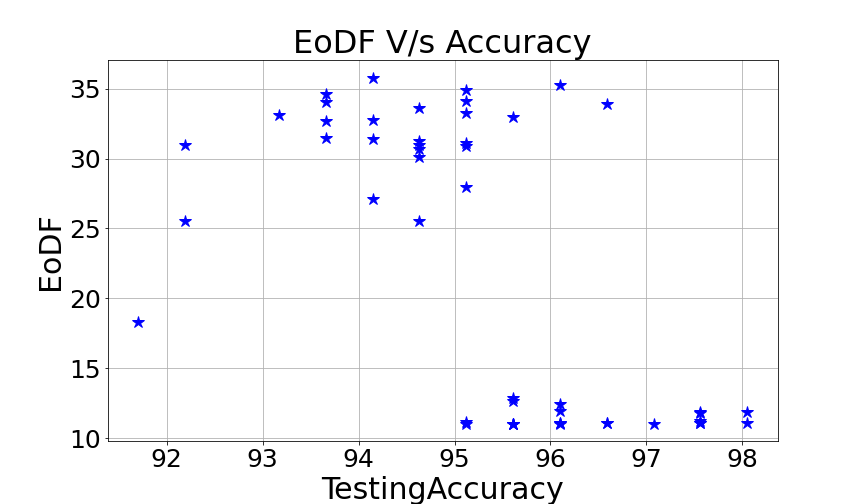}
\caption{Testing Accuracy}
\end{subfigure}
\caption{Effective degree of freedom (EoDF) V/s Accuracy for Cancer-10 dataset. The best $(K_i, q_i)$ parameter setting for this dataset is found to be $K^* = [ 8 , 3 , 8, 13,  8,  8, 13,  3,  8, 21]$ and $q^* = [2, 2, 5, 5, 2, 3, 3, 2, 2, 2]$.}
\label{fig:gam_cancer_10}
\end{figure}

\begin{table}[hbtp]
    \centering
    \caption{Details regarding parametric study for GAMs.}
    \label{tab:gam_parameters}
    \begin{tabular}{|c|c|}\hline
    \textbf{\# Basis Functions ($K$)}     & \textbf{Degree ($q$)}  \\\hline
     2, 3, 5, 8, 13, 21 &  2, 3, 5\\\hline
    \end{tabular}
\end{table}
Total number of terms arising from the expression of rule $g(\mathbf{x})$ (Eq.~\ref{eq:gam_splines}) is
$\sum_{j=1}^{d} {(q_j + 1)\times K_j + K_j + 1 }$.
However, due to second-order smoothening effect (Eq.~\ref{eq:gam_optimization}), 2nd order nonlinearities which are not contributing in minimizing the error $\sum_{i = 1}^N(y_i - \hat{y}_i(\mathbf{B'}\boldsymbol{\beta}))^2$ will get removed from the rule and thus, the \emph{effective degree of freedom} (EoDF) will be far less than the total length of the rule. 
Effective degrees of freedom versus accuracy plot for GAM classifiers obtained using various combinations of $(K_i, q_i)$ on Cancer-10 dataset is shown in Figure~\ref{fig:gam_cancer_10}. It is clear that a high training accuracy is achieved with a large EoDF, but makes an over-fitting and produces less testing accuracy. About 500 such experiments are performed and the best combinations of $(K_i, q_i)$ are used to generate results (Table~\ref{tab:customized_data_results}) for a given dataset. Note  here that generating classifiers using GAM is computationally expensive for high-dimensional datasets and so, we do not run experiments on datasets involving 500 features.


\noindent {\bf Advantages:}
\begin{itemize}
    \item Effect of each feature on the output variable can be separately analyzed using partial dependence plots. 
    \item A source code is available \cite{serven2018pygam} for rapid prototyping. 
\end{itemize}

\noindent {\bf Disadvantages:}
\begin{itemize}
\item Hyperparameters defining the non-linear functions Eq.~\ref{eq:gams} needs to be properly identified.
\item Slow to train as compared to other methods.
\item Becomes computationally expensive to handle high dimensional datasets.
\end{itemize}

\subsection{Genetic Programming (GP)}
\begin{wrapfigure}[8]{r}{0.47\linewidth}
\vspace{-12mm}
    \centering
    \includegraphics[width = \linewidth]{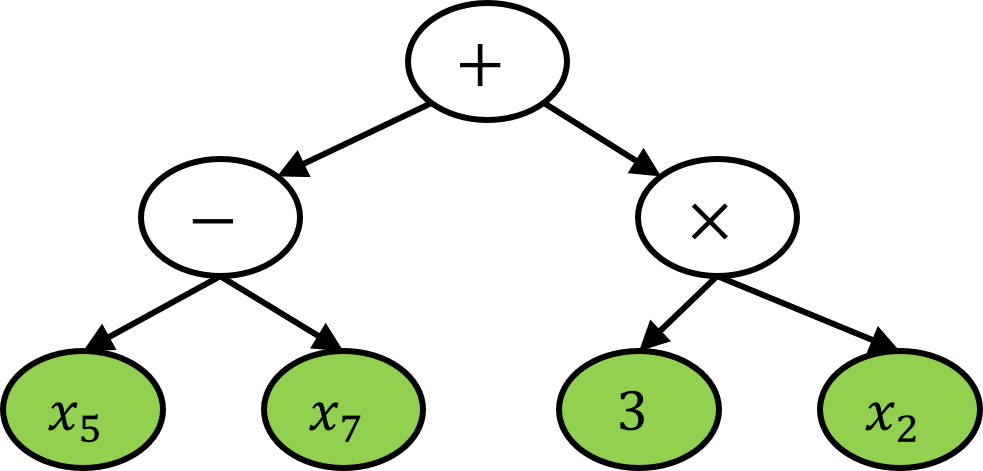}
    \caption{A sample genetic program (GP) tree. The above GP translates this equation: $f(\mathbf{x}) = (x_5 - x_7) + 3x_2$.}
    \label{fig:gp_tree_sample}
\end{wrapfigure}
Genetic Programming has been extensively used to derive non-linear and interpretable classifiers \cite{cano2013interpretable, de2002discovering, bot2000application, eggermont2004genetic, tan2002mining}. A GP algorithm evolves \emph{programs} (or equations of classifier's decision boundary in our case) using genetic operators like crossover and mutation. Programs in GP are usually represented with tree architecture as shown in Figure~\ref{fig:gp_tree_sample}. 
Internal nodes of this tree can involve mathematical operations, like $+, \times, -, \div, \log, \sin$. Allowable set of mathematical operations are pre-specified by the user. In our case, we use $\{+, \times, -, \div\}$ only. Terminal leaf nodes of a GP program either have one of the input feature $x_i$ or a constant term $c$. It is to note here that a GP tree ($\mathbf{T}$) represents one non-linear equation and is fundamentally different from the decision tree which involves assembly of split-rule equations which are organized in a hierarchical format (Figure~\ref{fig:cart_illustration}). The optimal structure of tree, operators used, features $x_i$ involved and value of constants $c$ are all unknown and are determined through an evolutionary algorithm.
The evolution is conducted with an objective to minimize the cross-entropy loss. However, if unchecked, the size of  GP trees grows as the evolution progress and the GP algorithm suffers from \emph{bloating} \cite{iba1999bagging}.
To counter this effect of bloating and encourage evolution of simpler trees (trees with less number of nodes), a parsimony coefficient $P_c$ is used to penalize the fitness of a GP tree ($\mathbf{T}$) as shown below:
\begin{equation}
    \begin{array}{rl}
         \textit{Min:} & f_{GP}(\mathbf{T}) = C_{loss} + P_c\times T_{size},
    \end{array}
    \label{eq:gp_optim}
\end{equation}
where,
\begin{equation}
\begin{array}{l}
    C_{loss}\!=\!-\mathlarger{\frac{1}{N}} \mathlarger{\sum}_{i=1}^{N}y(\mathbf{x}_i)\log(\hat{y}(\mathbf{x}_i))\!-\!(1\!-\!y(\mathbf{x}_i))\log(1\!-\! \hat{y}(\mathbf{x}_i)),\\
    \hat{y}(\mathbf{x})\!=\! \textrm{Sigmoid}(f(\mathbf{x})).
    \end{array}
\label{eq:sigmoid}
\end{equation}
In Eq.~\ref{eq:gp_optim}, $T_{size}$ represents size of the tree and is computed by counting total number of nodes in the tree. In Eq.~\ref{eq:sigmoid}, $f(\mathbf{x})$ is the value the GP tree outputs for a given feature vector $\mathbf{x}$ (see Figure~\ref{fig:gp_tree_sample}).

\begin{figure}[hbtp]
    \centering
    \begin{subfigure}[b]{0.52\linewidth}
    \includegraphics[width = \linewidth]{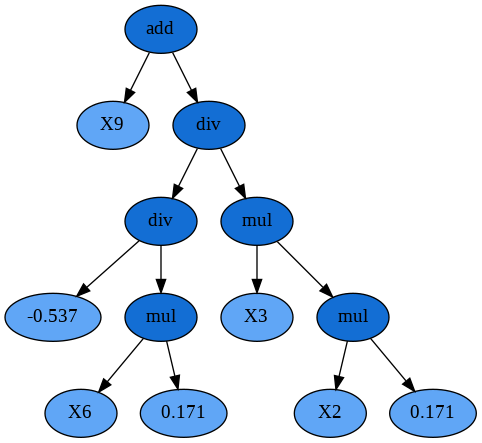}
    \caption{$P_c = 0.005$, $T_r = 96.44$, $T_e~=~99.02$, Complexity = 6.}
    \end{subfigure}%
    ~
    \begin{subfigure}[b]{0.46\linewidth}
    \includegraphics[width = 0.7\linewidth]{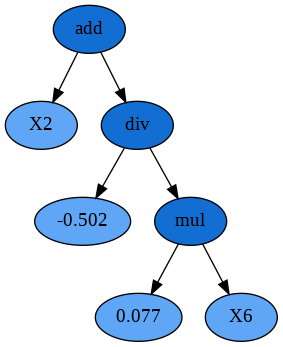}
    \caption{$P_c= 0.01$, $T_r = 95.60$, $T_e =  98.05$, Complexity = 3.}
    \end{subfigure}
    \caption{Classifiers for Cancer data:
    $P_c=0.005$: $f(\boldx) =  x_9 + \frac{-0.537}{(0.171x_6)(0.171x_3x_2)}$ and $P_c=0.01$: $f(\boldx) = x_2 + \frac{-0.502}{(0.077x_6)}$.}
    \label{fig:GP_Cancer_rule}
    \vspace{-15pt}
    
\end{figure}

It is important to choose a suitable parsimony coefficient $P_c$ for a problem. Smaller value of $P_c$ will encourage bloating and will evolve complex equations while the higher value of $P_c$ will evolve simpler equations at an expense of reduced classification accuracy. 
In our case, we perform experiments using three values $P_c$: 0.01, 0.005 and 0.001, and conduct 50 runs on each dataset shown in Table~\ref{tab:gp_results} (discussed in Section~\ref{sec:problems}) after randomly splitting the dataset into 70\% training and 30\% testing for each run. Statistics regarding testing accuracy and {\em complexity} (measured as the total number of internal nodes) is reported in the table.
\begin{table*}[hbt]
    \centering
    \caption{GP Result for different values of parsimony coefficient $P_c$. For each dataset, the first row represents the testing accuracy and the second row represents complexity (number of internal nodes). $P_c=0.001$ produces better results.}
    \label{tab:gp_results}
    \begin{tabular}{|c|c|c|c|c|c|c|}\hline
    \textbf{Pars. coeff.} & \textbf{DS1} & \textbf{DS2} & \textbf{DS3} & \textbf{DS4} & \textbf{Truss} & \textbf{WeldedBeam}\\\hline
    ${P_C = 0.01}$ & 
    
    \begin{tabular}{c}
         $61.07 \pm 9.91$\\	$\mathbf{3.40 \pm 3.70}$ 
    \end{tabular} & 
    \begin{tabular}{c}
        $95.24 \pm 0.00$ \\	$\mathbf{1.98 \pm 0.14}$ 
    \end{tabular} & \begin{tabular}{c}
        $65.37 \pm 11.57$\\	$\mathbf{4.44 \pm 2.37}$ 
    \end{tabular} & 
    \begin{tabular}{c}
       $49.93 \pm 1.43$ \\	$\mathbf{1.12 \pm 3.40}$ 
    \end{tabular} & \begin{tabular}{c}
        $82.78 \pm 11.28$\\	$\mathbf{5.20 \pm 3.30}$ 
    \end{tabular} & \begin{tabular}{c}
         $84.88 \pm 13.08$ \\	$\mathbf{9.32 \pm 5.15}$ 
    \end{tabular}\\\hline
    ${P_C = 0.005}$ & 
    \begin{tabular}{c}
        $77.3 \pm 11.29$  \\ $16.18 \pm 9.99$ 
    \end{tabular} & \begin{tabular}{c}
         $95.24 \pm 0.00$  \\ $3.86 \pm 1.23$ 
    \end{tabular} & \begin{tabular}{c}
         $86.27 \pm 11.41$  \\ $19.86 \pm 11.45$ 
    \end{tabular} & 
    \begin{tabular}{c}
        $50.37 \pm 2.96$  \\ $2.06 \pm 3.88$ 
    \end{tabular} & \begin{tabular}{c}
         $90.03 \pm 8.50$  \\ $11.98 \pm 7.12$ 
    \end{tabular} & \begin{tabular}{c}
         $92.35 \pm 6.06$  \\ $14.08 \pm 5.35$ 
    \end{tabular}\\\hline
    ${P_C = 0.001}$ & 
    \begin{tabular}{c}
        $\mathbf{91.70 \pm 6.91}$  \\$67.72 \pm 26.72$ 
    \end{tabular} & \begin{tabular}{c}
         $\mathbf{95.37 \pm 0.63}$  \\ $15.14 \pm 13.55$ 
    \end{tabular} & \begin{tabular}{c}
         $\mathbf{96.50 \pm 3.3}$  \\ $76.74 \pm 33.36$ 
    \end{tabular} & 
    \begin{tabular}{c}
        $\mathbf{58.00 \pm 11.22}$  \\ $18.76 \pm 23.94$ 
    \end{tabular} & \begin{tabular}{c}
         $\mathbf{97.36 \pm 3.81}$  \\ $36.02 \pm 16.99$ 
    \end{tabular} & \begin{tabular}{c}
         $\mathbf{96.46 \pm 4.14}$  \\ $35.90 \pm 18.28$ 
    \end{tabular}\\\hline
    \end{tabular}%
    \vspace{5pt}
    \begin{tabular}{|c|c|c|c|c|cH|}\hline
    \textbf{Pars. coeff.} & \textbf{m-DS1} & \textbf{m-DS2} & \textbf{m-DS3} & \textbf{Cancer-10} & \textbf{Cancer-30} & \textbf{\seethis{m-ZDT-500}}\\\hline
    ${P_C = 0.01}$ & 
    \begin{tabular}{c}
$89.53 \pm 3.27$ \\	$\mathbf{8.34 \pm 1.98}$
    \end{tabular} & \begin{tabular}{c}
       $95.65 \pm 0.70$\\	$\mathbf{3.58 \pm 1.07}$
    \end{tabular} & \begin{tabular}{c}
       $96.33 \pm 4.68$ \\$\mathbf{15.04 \pm 6.36}$
    \end{tabular}  & \begin{tabular}{c}
        $94.03 \pm 4.59$ \\ $\mathbf{5.56 \pm 2.06}$
    \end{tabular} & \begin{tabular}{c}
        $90.47 \pm 4.54$ \\	$\mathbf{4.78 \pm 2.30}$ 
    \end{tabular} & 
    \begin{tabular}{c}
        \textbf{--}  \\ comp
    \end{tabular}\\\hline
    ${P_C = 0.005}$ & 
    \begin{tabular}{c}
        $93.37 \pm 4.57$  \\ $16.32 \pm 9.55$ 
    \end{tabular} & \begin{tabular}{c}
         $95.65 \pm 0.70$  \\ $3.76 \pm 1.22$ 
    \end{tabular} & \begin{tabular}{c}
         $98.4 \pm 1.99$  \\ $19.88 \pm 9.94$ 
    \end{tabular} & \begin{tabular}{c}
         $95.04 \pm 1.76$  \\ $7.88 \pm 3.07$ 
    \end{tabular} & \begin{tabular}{c}
         $90.96 \pm 6.29$  \\ $5.74 \pm 1.84$ 
    \end{tabular}  & 
    \begin{tabular}{c}
        \textbf{--} \\ comp 
    \end{tabular}\\\hline
    ${P_C = 0.001}$ & 
    \begin{tabular}{c}
        $\mathbf{98.83 \pm 1.88}$  \\ $55.38 \pm 22.39$ 
    \end{tabular} & \begin{tabular}{c}
         $\mathbf{96.67 \pm 1.93}$  \\ $14.08 \pm 9.11$ 
    \end{tabular} & \begin{tabular}{c}
         $\mathbf{99.27 \pm 1.22}$  \\ $49.80 \pm 21.69$ 
    \end{tabular} & \begin{tabular}{c}
         $\mathbf{96.13 \pm 1.29}$	\\ $15.80 \pm 5.66$
    \end{tabular} & \begin{tabular}{c}
        $\mathbf{92.40 \pm 4.98}$ \\ $14.58 \pm 7.14$
    \end{tabular} & 
    \begin{tabular}{c}
        \textbf{--}  \\ comp 
    \end{tabular}\\\hline
    \end{tabular}
\end{table*}
It is clear from the table that while a small $P_c$ produces a better accuracy, a large $P_c$ produces smaller sized GPs. 
To demonstrate, we present two GP classifiers for $P_c=0.005$ and $0.01$ obtained  for the breast cancer Wisconsin dataset (involving total 10 features) in Figure~\ref{fig:GP_Cancer_rule}. Training ($T_r$) and testing ($T_e$) accuracy are better for $P_c=0.005$. 


Table~\ref{tab:gp_results} indicates that GP does not perform well on certain problems even in small-sized problems, such as DS1 and DS4. In a mathematical classifier search, there are two hierarchical aspects which must be learnt: (i) structure of the classifier, and (ii) coefficient of each term in the structure. GP attempts to learn both aspects in a single optimization task. We argue that while a ``good" structure may have evolved at a generation, if its associated coefficients are not proper, the whole classifier will be judged as ``bad". We attempt to alleviate this aspect in the next procedure by using a bilevel optimization framework.  

\pagebreak
\noindent {\bf Advantages:}
\begin{itemize}
    \item Non-linearity gets automatically determined during evolution.
    \item Open Source Code is available \url{https://gplearn.readthedocs.io/en/stable/index.html}.
\end{itemize}

\noindent {\bf Disadvantages:}
\begin{itemize}
    \item Correct set of operators needs to be specified to derive optimal interpretable classifier.
    \item Training is slow as compared to SVM and CART.
    \item Parsimony coefficient $P_c$  severely impacts the performance of GP and so it needs to be tuned properly. 
\end{itemize}
\vspace{-4pt}
\subsection{Nonlinear Decision Tree (NLDT) Approach}
\label{sec:nldt}
\begin{figure}[hbtp]
    \centering
    \includegraphics[width = 0.9\linewidth]{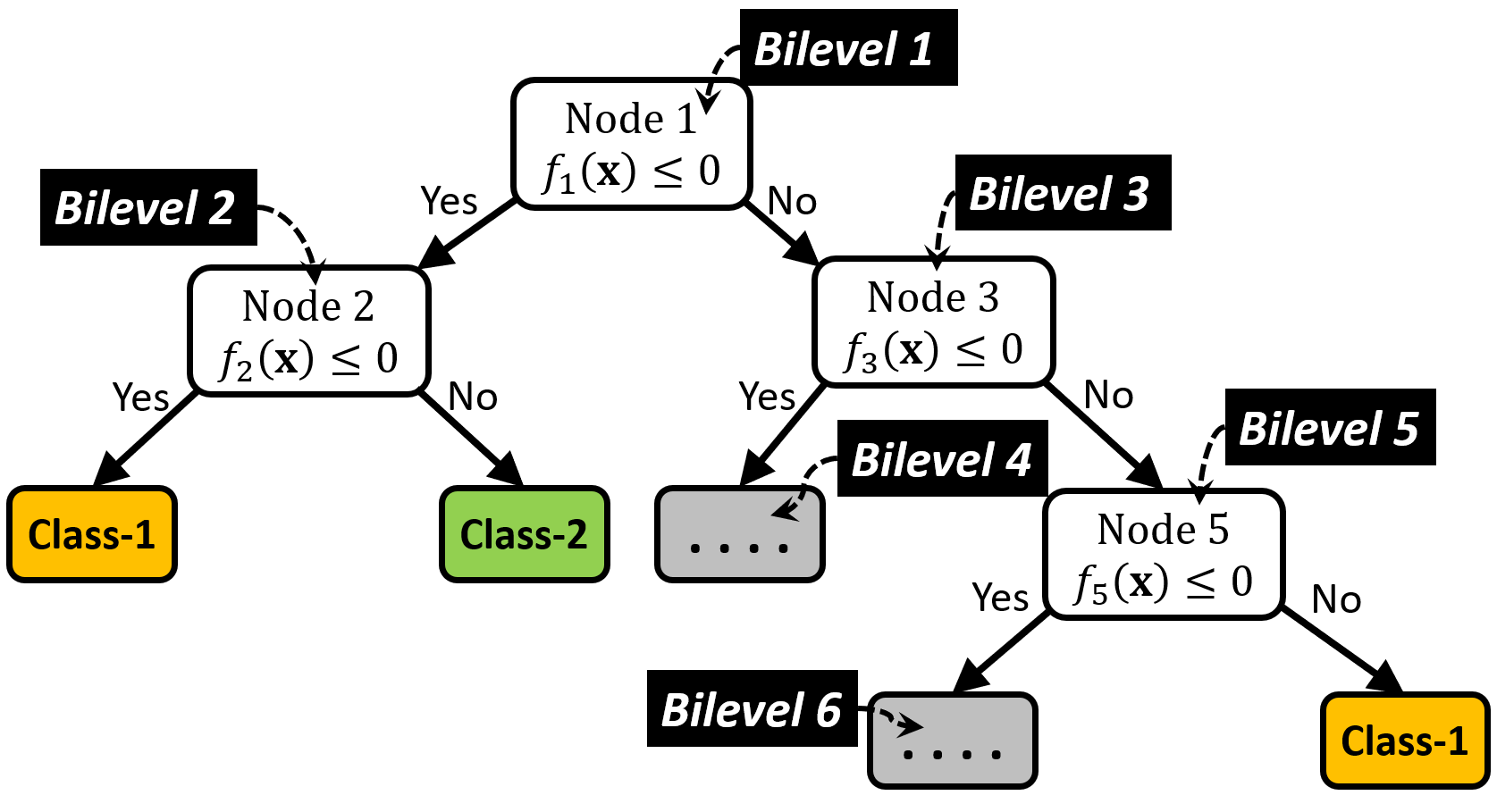}
    \caption{NLDT Schematic.}
    \label{fig:nldt_schematic}
\end{figure}

Recently, an evolutionary algorithm based non-linear decision tree classifier was proposed in \cite{dhebar2020interpretable}. The classifier is represented in the form of a \emph{non-linear} decision tree as shown in Figure~\ref{fig:nldt_schematic}. 
Unlike in regular CART based decision tree where the split-functions are constrained to have axis-parallel structure (Figure~\ref{fig:cart_illustration}), split-functions $f_i(\mathbf{x})$ in NLDT are non-linear to the features and are represented as weighted sum of $p$ power-laws as shown below:
\begin{align}
    f(\mathbf{x}) = \begin{cases}
    \mathlarger{\sum}_{i=1}^pw_iB_i + \theta_1, \quad \mbox{if $m = 0$},\\
    \left|\mathlarger{\sum}_{i=1}^pw_iB_i + \theta_1\right| - \theta_2, \quad \mbox{if $m = 1$},
    \end{cases}
    \label{eq:nldt_split_rule}
\end{align}
where $B_i$ are the power-laws ($B_i = \prod_{j = 1}^d x_j^{b_{ij}}$), $w_i$ are coefficients, $\theta_i$ are biases, and $d$ is the number of features in  the dataset.
The exponents $b_{ij}$ of the $j$-th feature in the $i$-th power-law can assume a value from a pre-specified discrete set $E$. In our case, we choose $E = {-1, -2, \dots, 3}$. The number of power-laws $p$ is set to 3 in all the experiments. At each conditional node in NLDT, the expression for split-rule $f(\mathbf{x})$ is derived by optimizing exponents $b_{ij}$, coefficients $w_{i}$, biases $\theta_i$ and the modulus-flag $m$ using a dedicated bilevel algorithm as shown in Figure~\ref{fig:nldt_schematic}. The upper level of the bilevel algorithm operates in the discrete space of exponents $b_{ij}$ (which are encoded using a matrix $\mathbf{B}$) and the modulus flag $m$ while for each upper level solution $S_U$, the lower level algorithm searches for the optimal values of weights $\mathbf{w}$ and biases $\mathbf{\Theta}$. The upper level is modeled as a single-objective constrained optimization problem with an objective to minimize the complexity $F_U$ of the split-rule $f(\mathbf{x})$ while ensuring that child nodes resulting from split have their net impurity $F_L$ less than a user specified threshold value $\tau_I$ (set to 0.05 in our experiments). The bilevel optimization formulation to derive a split-rule $f(\mathbf{x})$ in NLDT can then be written as shown below:
\begin{equation}
\hspace{-1ex}\begin{array}{rl}
\text{Min.}&   
    F_U(\mathbf{B},m,\mathbf{w}^{\ast}, \mathbf{\Theta}^{\ast}), \\
    \text{s.t.} & 
 (\mathbf{w}^{\ast}, \mathbf{\Theta}^{\ast})\!\in\!{\rm argmin}\!\left\{F_L(\mathbf{w}, \mathbf{\Theta})|_{(\mathbf{B}, m)} \big| F_L(\mathbf{w}, \mathbf{\Theta})|_{(\mathbf{B},m)}\right. \\ 
 & \left.\quad \leq \tau_I,
 -1\leq w_i \leq 1, \ \forall i,\ \mathbf{\Theta} \in [-1,1]^{m+1}\right\}, \\
 & m \in \{0,1\},\ b_{ij} \in \{-3,-2,-1,0,1,2,3\}.
\end{array}
\label{eq:bilevel_formulation_binary_split}
\end{equation}

\begin{figure}[hbt]
    \centering
    \includegraphics[width = \linewidth]{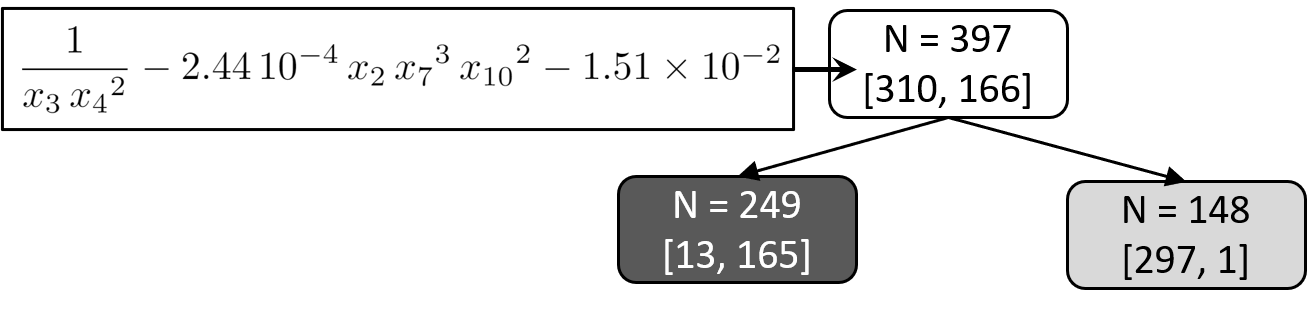}
    \caption{NLDT with a single nonlinear rule obtained for Breast Cancer Wisconsin dataset (total 10 features) is shown. If the function value on the root node is less-than-equal-to zero, the data point is classified as Class 2 point with 165/178 or 92.7\% accuracy and if it is positive, then the data point is classified as Class 1 point with 297/298 or 99.7\% accuracy.}
    \label{fig:nldt_cancer_10}
\end{figure}

The upper level objective $F_U$, which quantifies the complexity is computed by counting all non-zero exponents $b_{ij}$ in the expression of $f(\mathbf{x})$ (Eq.~\ref{eq:nldt_split_rule}). The lower level objective function $F_L$ which quantifies the quality of split is obtained using the weighed sum of impurities of child nodes as shown in Eq.~\ref{eq:split_quality}.
Evolutionary algorithms for both upper and lower level are employed to conduct an efficient search on upper level variables ($\mathbf{B}$, $m$) and lower level variables ($\mathbf{w}$, $\mathbf{\Theta}$). Splits in NLDT are recursively derived until a certain termination criteria is met. The bilevel optimization serves as a very efficient search technique to derive simple split rules, an example of which is shown in Figure~\ref{fig:nldt_cancer_10} for Wisconsin breast cancer dataset involving total 10 features. Besides being an interpretable classifier, it also reveals that only five ($x_2$-$x_4$, $x_7$ and $x_{10}$) of ten features are important in making the classification. 
\vspace{-3mm}

\noindent {\bf Advantages:}
\begin{itemize}
    \item Due to the use of nonlinear structure, the NLDT will have a fewer rules,
    \item The structure of the rules can be controlled easily, so interpretable rules can be obtained.
    \item Recent advancements in nonlinear optimization methods enable NLDTs to be evolved efficiently. 
\end{itemize}

\noindent {\bf Disadvantages:}
\begin{itemize}
    \item Maximum depth, total number of power laws per rule, exponent set, impurity threshold $\tau_I$ and minimum number of data points to conduct split, need to be set. 
    \item Training is slower as compared to CART and SVM.
\end{itemize}

Details regarding the bilevel optimization algorithm and parameter settings can be found from \cite{dhebar2020interpretable}.

\section{Datasets Considered} \label{sec:problems}

In our study, we conduct experiments on total 19 datasets to explore and investigate behaviour of various classification algorithms discussed above on varieties of features spaces and data distributions. 

\subsection{Customized Data: DS1-4 and modified DS1-3}
\begin{figure}[hbt]
    \begin{subfigure}{0.49\linewidth}
    \centering\includegraphics[width = 1.1\linewidth]{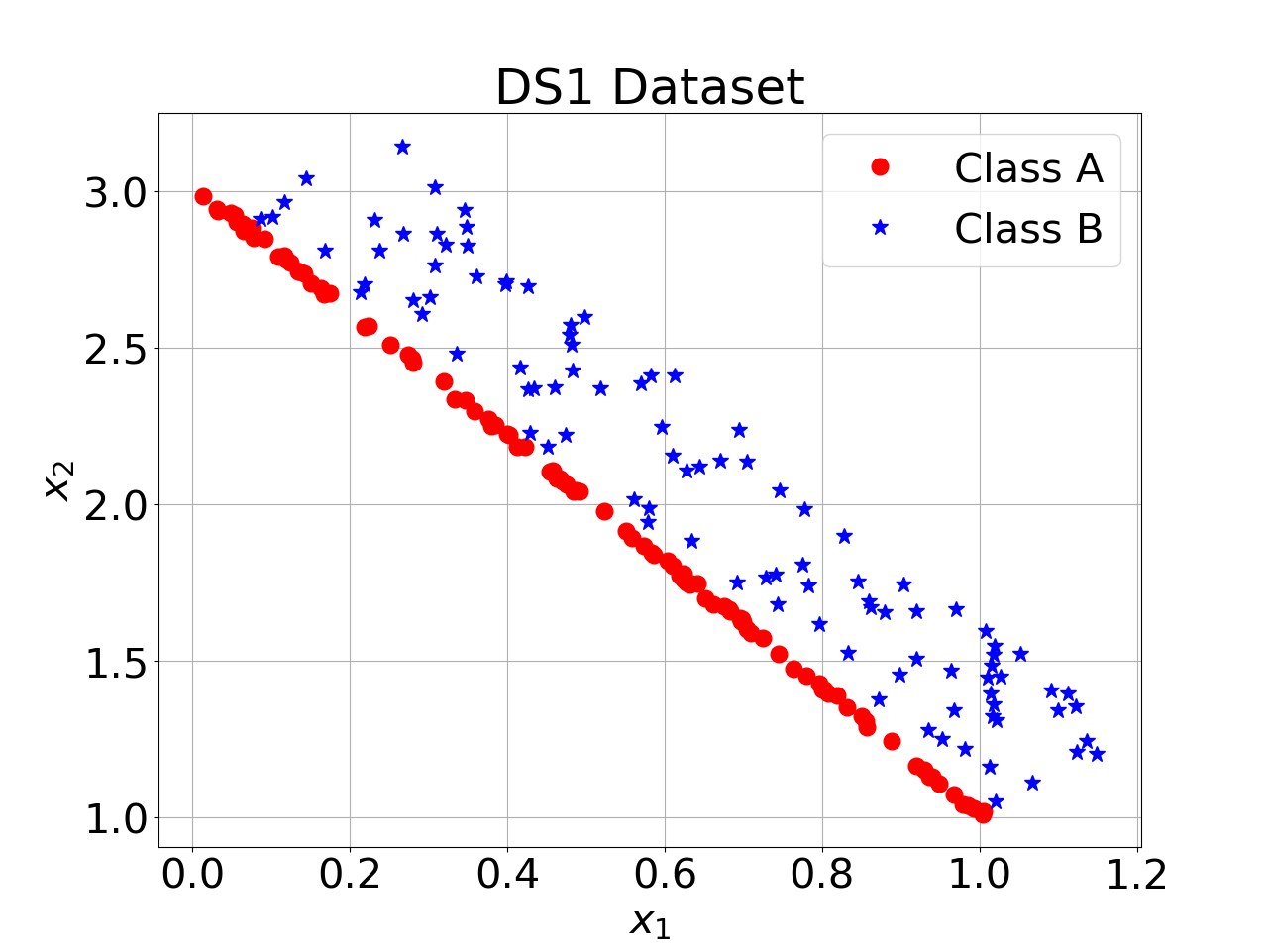}
    \caption{DS1 Dataset.}
    \label{fig:DS1_dataset}
\end{subfigure}
\begin{subfigure}{0.49\linewidth}
    \centering
    \includegraphics[width = 1.1\linewidth]{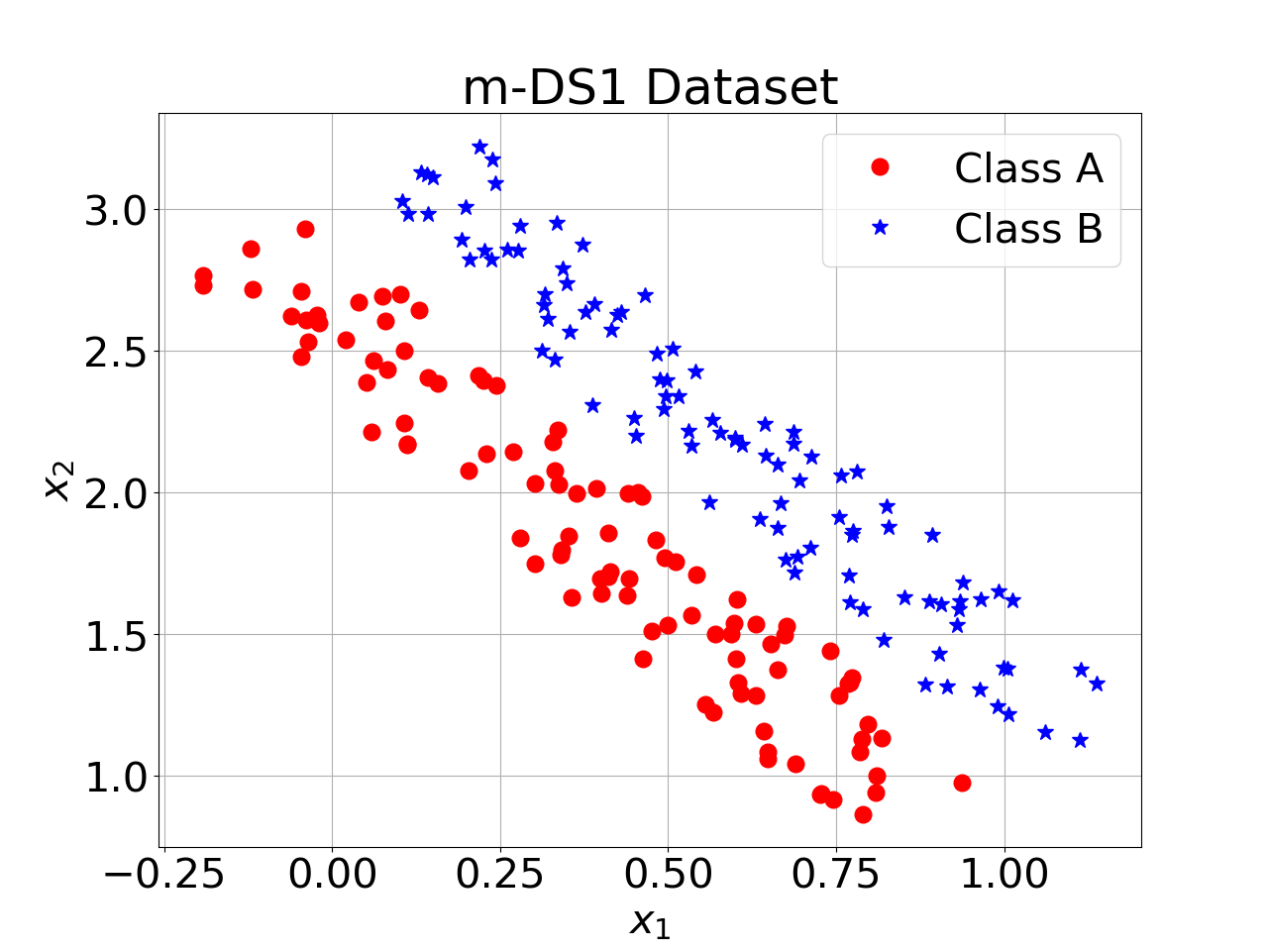}
    \caption{m-DS1 Dataset.}
    \label{fig:modified_DS1_dataset}
\end{subfigure}
\caption{Original DS1 and its modified version.}
\label{fig:DS1}
\end{figure}

Four synthetic two dimensional datasets DS1-DS4 and their variants m-DS1, m-DS2 and m-DS3 are generated using the procedure provied in \cite{dhebar2020interpretable}  to investigate behavior of classification algorithms across following properties:
\begin{itemize}
    \item \textbf{Data Distribution:} For DS1-DS4 datasets, degree of scatter in data varies across classes. For m-DS1, m-DS2 and m-DS3 the scattering of data for each class is more similar than that in original DS datasets. A visualisation of feature spaces for DS1 and m-DS1 dataset is provided in Figure~\ref{fig:DS1_dataset} and \ref{fig:modified_DS1_dataset}, respectively.
    \item \textbf{Geometry of Decision Boundary:} Here, the effect of the nature of the simplest possible decision boundary is considered. Decision boundary corresponding to DS1-DS2 and modified DS1-DS2 is linear, DS3 and m-DS3 have decision boundary involving nonlinearity of order 2 and DS4 have two disjoint linear decision boundaries. 
    
    \item \textbf{Data Bias:} Here, effect of bias in class representation is considered. All datasets except DS2 and m-DS2 are balanced. For DS2 and m-DS2, minority class has 5 times less number of data points as the majority class.
\end{itemize}

\subsection{Cancer Datasets}
We use breast cancer Wisconsin data involving 10 features and Wisconsin Diagnostics dataset having 30 features. 

\subsection{Pareto versus Non-Pareto Classification}
In multi-objective optimization, there are two types of solutions: (i) Pareto-optimal set and (ii) Dominated set. Users are interested in knowing what feature relationships (decision variables interactions) make a solution Pareto-optimal, thereby making the task a binary classification problem. 

\subsubsection{Test problems}
We use modified versions of ZDT \cite{eckideb} and DTLZ \cite{deb-scalable} problems with two and three objectives, respectively to generate datasets involving 30 and 500 features (details in \cite{dhebar2020interpretable}). These two problem sizes also allow us to perform a scale-up study of the classification methods. 

\subsubsection{Real-world Problems}
Next, we consider two real-world problems -- welded beam and truss design problems \cite{deb-book-01}. 

\section{Results and Discussions}\label{sec:results}
Table~\ref{tab:customized_data_results} presents the testing accuracy and complexity of five classification methods on 19 problems. For each method, a parametric study is performed on each problem and the setting which obtained the best testing accuracy is used to generate the final results. Statistics of 50 runs (with random data split of 70\% training and 30\% testing in each) on each dataset for two performance metrics is presented in Table~\ref{tab:customized_data_results}.
\begin{table*}[hbt]
\caption{Summary of results obtained using various methods. For each dataset, the first row indicates testing accuracy and the second row indicates complexity. Italicized entries are statistically insignificant (according to 95\% confidence in Wilcoxon rank-sum test) compared to the best entry in the same row.} 
    \label{tab:customized_data_results}
    \centering
    \begin{tabular}{|c|cH|c|c|c|c|c|}\hline
        \textbf{Sr.} & \textbf{Problem} & \textbf{Property} & \textbf{NLDT}  &
        \textbf {CART} & \textbf{SVM} & \textbf{GAM} & \textbf{GP}\\\hline
         1 & DS1 & 
         \begin{tabular}{c}
              Accuracy \\ Complexity 
         \end{tabular} & 
         \begin{tabular}{c}
              $ { 99.55 \pm 1.08} $ \\ $\mathbf{2.3 \pm 0.6}$
         \end{tabular} & 
         \begin{tabular}{c}
              $90.32 \pm 4.06 $ \\ $14.5 \pm 1.7$ 
         \end{tabular} & 
         \begin{tabular}{c}
              $ 99.87 \pm 0.45 $ \\ $ 8.16 \pm 0.88 $ 
         \end{tabular} & 
         \begin{tabular}{c}
         
      $\mathbf{100.0 \pm 0.00}$	\\	$2.89 \pm 0.00$
            \end{tabular} & 
         \begin{tabular}{c}
            $91.70 \pm 6.91$	\\	$67.72 \pm 26.72$
         \end{tabular} \\\hline
         
         2 & DS2 & 
         \begin{tabular}{c}
              Accuracy \\ Complexity 
         \end{tabular} & 
         \begin{tabular}{c}
              ${ 99.44 \pm 0.87 }$ \\ $\mathbf{ 2.3 \pm 0.7}$
         \end{tabular} & 
         \begin{tabular}{c}
              $95.43 \pm 1.50$ \\ $11.0 \pm 1.4$ 
         \end{tabular} & 
         \begin{tabular}{c}
              $ 99.33 \pm 1.10 $ \\ $ 7.64 \pm 0.87 $
         \end{tabular} & 
         \begin{tabular}{c}
      $\mathbf{100.0 \pm 0.00}$	\\	$2.89 \pm 0.00$
             
         \end{tabular} & 
         \begin{tabular}{c}
              $95.37 \pm 0.63$	\\	$15.14 \pm 13.55$
         \end{tabular} \\\hline
         
         3 & DS3 & 
         \begin{tabular}{c}
              Accuracy \\ Complexity 
         \end{tabular} & 
         \begin{tabular}{c}
              $\mathbf{ 99.77 \pm 0.67} $ \\ $\mathbf{ 2.2 \pm 0.5} $
         \end{tabular} & 
         \begin{tabular}{c}
              $95.00 \pm 2.35 $ \\ $11.5 \pm 1.3$ 
         \end{tabular} & 
         \begin{tabular}{c}
              $\mathit{ 99.63 \pm 0.69} $ \\ $ 10.22 \pm 1.42 $
         \end{tabular} & 
         \begin{tabular}{c}
      $\mathit{99.47 \pm 1.03}$	\\	$4.98 \pm 0.14$
              
         \end{tabular} & 
         \begin{tabular}{c}
           $96.50 \pm 3.30$	\\	$76.74 \pm 33.36$
         \end{tabular} \\\hline
         
         4 & DS4 & 
         \begin{tabular}{c}
              Accuracy \\ Complexity 
         \end{tabular} & 
         \begin{tabular}{c}
              $\mathbf{98.88 \pm 1.65} $ \\ $\mathbf{3.1 \pm 1.4}$
         \end{tabular} & 
         \begin{tabular}{c}
              $88.68 \pm 3.60 $ \\ $31.3 \pm 4.2$ 
         \end{tabular} & 
         \begin{tabular}{c}
              $ 93.97 \pm 2.35 $ \\ $ 43.70 \pm 2.69 $
         \end{tabular} & 
         \begin{tabular}{c}
        $48.63 \pm 6.50$	\\	$3.80 \pm 0.99$
         \end{tabular} & 
         \begin{tabular}{c} 
             $59.63 \pm 10.81$ \\	$24.70 \pm 26.40$

         \end{tabular} \\\hline
         
         5 & m-DS1 & 
         \begin{tabular}{c}
              Accuracy \\ Complexity 
         \end{tabular} & 
         \begin{tabular}{c}
            $ 99.10 \pm 1.54 $ \\ $\mathbf{ 2.00 \pm 0.00 }$
         \end{tabular} & 
         \begin{tabular}{c}
              $ 89.73 \pm 4.53 $ \\ $ 7.90 \pm 1.22 $
         \end{tabular} & 
         \begin{tabular}{c}
              $ 99.90 \pm 0.40 $ \\ $ 7.50 \pm 0.75 $
         \end{tabular} & 
         \begin{tabular}{c}
        $\mathbf{100.0 \pm 0.00}$	\\	$2.90 \pm 0.00$              
         \end{tabular} & 
         \begin{tabular}{c}
            $98.83 \pm 1.88$	\\	$55.38 \pm 22.39$
         \end{tabular} \\\hline
         
         6 & m-DS2 & 
         \begin{tabular}{c}
              Accuracy \\ Complexity 
         \end{tabular} & 
         \begin{tabular}{c}
              $ 99.46 \pm 1.08 $ \\ $\mathbf{ 2.10 \pm 0.30 }$
         \end{tabular} & 
         \begin{tabular}{c}
              $ 96.25 \pm 1.92 $ \\ $ 5.96 \pm 0.81 $ 
         \end{tabular} & 
         \begin{tabular}{c}
              $ 99.94 \pm 0.44 $ \\ $ 5.44 \pm 0.67 $ 
         \end{tabular} & 
         \begin{tabular}{c}
        $\mathbf{99.94 \pm 0.31}$	\\	$2.90 \pm 0.00$             
         \end{tabular} & 
         \begin{tabular}{c}
              $96.67 \pm 1.93$	\\	$14.08 \pm 9.11$
         \end{tabular} \\\hline
         
         7 & m-DS3 & 
         \begin{tabular}{c}
              Accuracy \\ Complexity 
         \end{tabular} & 
         \begin{tabular}{c}
              $ 99.20 \pm 1.30 $ \\ $\mathbf{ 2.02 \pm 0.14 }$
         \end{tabular} & 
         \begin{tabular}{c}
              $ 92.87 \pm 4.35 $ \\ $ 5.78 \pm 1.11 $ 
         \end{tabular} & 
         \begin{tabular}{c}
         
             $\mathbf{ 100.00 \pm 0.00 }$	\\	$ 8.82 \pm 0.89 $
         \end{tabular} & 
         \begin{tabular}{c}
         $99.17 \pm 1.48$	\\	$3.24 \pm 0.22$
              
         \end{tabular} & 
         \begin{tabular}{c}
           ${99.27 \pm 1.22}$	\\	$49.8 \pm 21.69$
         \end{tabular} \\\hline\hline
         
         8 & Cancer-10 & 
         \begin{tabular}{c}
              Accuracy \\ Complexity 
         \end{tabular} & 
         \begin{tabular}{c}
              $\mathbf{96.50 \pm 1.16}$ \\ $\mathbf{ 6.4 \pm 1.7 }$
         \end{tabular} & 
         \begin{tabular}{c}
              $94.34 \pm 1.92 $ \\ $11.6 \pm 2.4$ 
         \end{tabular} & 
         \begin{tabular}{c}
              $ 95.07 \pm 1.23 $ \\ $ 51.26 \pm 5.02 $
         \end{tabular} & 
         \begin{tabular}{c}
        $95.32 \pm 1.49$	\\	$22.14 \pm 10.36$
              
         \end{tabular} & 
         \begin{tabular}{c}
             $96.13 \pm 1.29$	\\	$15.80 \pm 5.66$
         \end{tabular} \\\hline
         
         9 & Cancer-30 & 
         \begin{tabular}{c}
              Accuracy \\ Complexity 
         \end{tabular} & 
         \begin{tabular}{c}
              $\mathbf{96.20 \pm 1.49 }$ \\ $\mathbf{ 9.2 \pm 4.1}$
         \end{tabular} & 
         \begin{tabular}{c}
         
              $92.11 \pm 2.07 $ \\ $10.8 \pm 2.1$ 
         \end{tabular} & 
         \begin{tabular}{c}
            $ 95.24 \pm 1.29 $ \\ $ 58.88 \pm 4.46 $
         \end{tabular} & 
         \begin{tabular}{c}
            $93.74 \pm 5.83$	\\	$32.47 \pm 12.41$
         \end{tabular} & 
         \begin{tabular}{c}
             $92.40 \pm 4.98$	\\	$14.58 \pm 7.14$
         \end{tabular} \\\hline\hline
         
         10 & Welded Beam & 
         \begin{tabular}{c}
              Accuracy \\ Complexity 
         \end{tabular} & 
         \begin{tabular}{c}
              $98.58 \pm 1.13 $ \\ $\mathbf{ 3.9 \pm 1.0}$
         \end{tabular} & 
         \begin{tabular}{c}
              $97.72 \pm 1.04 $ \\ $8.42 \pm 1.42$ 
         \end{tabular} & 
         \begin{tabular}{c}
              $\mathbf{ 99.58 \pm 0.45 }$ \\ $ 7.86 \pm 1.27 $
         \end{tabular} & 
         \begin{tabular}{c}
         
       $\mathit{99.53 \pm 0.48}$	\\	$11.06 \pm 0.81$
         \end{tabular} & 
         \begin{tabular}{c}
              $96.46 \pm 4.14$	\\	$35.90 \pm 18.28$
         \end{tabular} \\\hline
         
         11 & Truss & 
         \begin{tabular}{c}
              Accuracy \\ Complexity 
         \end{tabular} & 
         \begin{tabular}{c}
              $\mathbf{ 99.54 \pm 0.75}$ \\ $\mathbf{ 3.30 \pm 0.90}$
         \end{tabular} & 
         \begin{tabular}{c}
              $98.33 \pm 1.10 $ \\ $11.06 \pm 3.15$ 
         \end{tabular} & 
         \begin{tabular}{c}
              $ 88.21 \pm 1.62 $ \\ $ 174.28 \pm 8.49 $ 
         \end{tabular} & 
         \begin{tabular}{c}
         $96.18 \pm 1.20$	\\	$19.19 \pm 1.06$
         \end{tabular} & 
         \begin{tabular}{c}
           $97.36 \pm 3.81$	\\	$36.02 \pm 16.99$
         \end{tabular} \\\hline\hline
         
         12 & m-ZDT1-30 & 
         \begin{tabular}{c}
              Accuracy \\ Complexity 
         \end{tabular} & 
         \begin{tabular}{c}
              ${98.97 \pm 0.57}$ \\ $\mathbf{7.60 \pm 3.50}$
         \end{tabular} & 
         \begin{tabular}{c}
              $ 97.77 \pm 0.58 $ \\ $ 30.26 \pm 4.65 $ 
         \end{tabular} & 
         \begin{tabular}{c}
              $\mathbf{ 99.39 \pm 0.35 }$ \\ $ 82.08 \pm 4.19 $
         \end{tabular} & 
         \begin{tabular}{c}
           $85.31 \pm 1.35$	\\	$220.20 \pm 11.73$
         \end{tabular} & 
         \begin{tabular}{c}
              $93.58 \pm 10.21$	\\	$45.34 \pm 26.09$
         \end{tabular} \\\hline
         
          13 & m-ZDT1-500 & 
         \begin{tabular}{c}
              Accuracy \\ Complexity 
         \end{tabular} & 
         \begin{tabular}{c}
              $ 98.93 \pm 0.60 $ \\ $\mathbf{9.34 \pm 4.15}$
         \end{tabular} & 
         \begin{tabular}{c}
              $ 95.96 \pm 0.80 $ \\ $ 21.02 \pm 1.55 $ 
         \end{tabular} & 
         \begin{tabular}{c}
              $\mathbf{100.00 \pm 0.00}$ \\ $ 140.58 \pm 4.25 $
         \end{tabular} & 
         \begin{tabular}{c}
             \textbf{---} \\ \textbf{---}
         \end{tabular} & 
         \begin{tabular}{c}
             $83.21 \pm 18.42$	\\	$52.14 \pm 24.47$
         \end{tabular} \\\hline
         
         14 & m-ZDT2-30 & 
         \begin{tabular}{c}
              Accuracy \\ Complexity 
         \end{tabular} & 
         \begin{tabular}{c}
              $ 98.96 \pm 0.57 $ \\ $\mathbf{8.10 \pm 3.35}$
         \end{tabular} & 
         \begin{tabular}{c}
              $ 97.88 \pm 0.70 $ \\ $ 28.22 \pm 2.35 $ 
         \end{tabular} & 
         \begin{tabular}{c}
              $\mathbf{99.51 \pm 0.33}$ \\ $ 80.98 \pm 3.50 $
         \end{tabular} & 
         \begin{tabular}{c}
            $84.97 \pm 1.18$	\\	$233.69 \pm 6.56$
         \end{tabular} & 
         \begin{tabular}{c}
              $91.57 \pm 11.92$	\\	$48.44 \pm 23.69$
         \end{tabular} \\\hline
         
         15 & m-ZDT2-500 & 
         \begin{tabular}{c}
              Accuracy \\ Complexity 
         \end{tabular} & 
         \begin{tabular}{c}
              $ 98.87 \pm 0.72 $ \\ $\mathbf{8.84 \pm 3.95}$
         \end{tabular} & 
         \begin{tabular}{c}
              $ 95.96 \pm 0.80 $ \\ $ 21.02 \pm 1.55 $
         \end{tabular} & 
         \begin{tabular}{c}
              $\mathbf{100.00 \pm 0.00}$ \\ $ 140.56 \pm 4.00 $
         \end{tabular} & 
         \begin{tabular}{c}
              \textbf{---} \\ \textbf{---}
         \end{tabular} & 
         \begin{tabular}{c}
         $85.06 \pm 16.82$	\\	$50.64 \pm 21.24$
         \end{tabular} \\\hline
         
         16 & m-DTLZ1-30 & 
         \begin{tabular}{c}
              Accuracy \\ Complexity 
         \end{tabular} & 
         \begin{tabular}{c}
              $\mathbf{98.77 \pm 0.87}$ \\ $\mathbf{11.98 \pm 5.85}$
         \end{tabular} & 
         \begin{tabular}{c}
              $ 78.52 \pm 7.94 $ \\ $ 128.40 \pm 22.39 $ 
         \end{tabular} & 
         \begin{tabular}{c}
              $ 94.22 \pm 0.95 $ \\ $ 615.54 \pm 9.24 $
         \end{tabular} & 
         \begin{tabular}{c}
             $55.96 \pm 3.19$	\\	$33.89 \pm 0.01$
         \end{tabular} & 
         \begin{tabular}{c}
          $81.59 \pm 17.32$	\\	$16.08 \pm 21.89$
         \end{tabular} \\\hline
         
         17 & m-DTLZ1-500 & 
         \begin{tabular}{c}
              Accuracy \\ Complexity 
         \end{tabular} & 
         \begin{tabular}{c}
              $\mathbf{93.76 \pm 4.24}$ \\ $22.72 \pm 7.50 $
         \end{tabular} & 
         \begin{tabular}{c}
              $ 78.31 \pm 7.21 $ \\ $ 126.94 \pm 20.07 $
         \end{tabular} & 
         \begin{tabular}{c}
              $ 64.32 \pm 1.76 $ \\ $ 1236.82 \pm 13.26 $ 
         \end{tabular} & 
         \begin{tabular}{c}
              \textbf{---}  \\ \textbf{---}  
         \end{tabular} & 
         \begin{tabular}{c}
           $80.49 \pm 11.54$	\\	$\mathbf{8.66 \pm 16.19}$
         \end{tabular} \\\hline
         
         18 & m-DTLZ2-30 & 
         \begin{tabular}{c}
              Accuracy \\ Complexity 
         \end{tabular} & 
         \begin{tabular}{c}
             $ \mathbf{ 97.22 \pm 2.25} $ \\ $17.48 \pm 6.75$
         \end{tabular} & 
         \begin{tabular}{c}
              $ 69.83 \pm 6.16 $ \\ $ 156.00 \pm 16.09 $ 
         \end{tabular} & 
         \begin{tabular}{c}
              $ 94.25 \pm 1.04 $ \\ $ 615.44 \pm 10.67 $ 
         \end{tabular} & 
         \begin{tabular}{c}
           $54.52 \pm 2.96$	\\	$35.84 \pm 0.02$
         \end{tabular} & 
         \begin{tabular}{c}
             $79.81 \pm 19.46$	\\	$\mathbf{12.32 \pm 14.79}$
         \end{tabular} \\\hline
         
         19 & m-DTLZ2-500 & 
         \begin{tabular}{c}
              Accuracy \\ Complexity 
         \end{tabular} & 
         \begin{tabular}{c}
              $ \mathbf{95.32 \pm 4.45} $ \\ $ 22.02 \pm 10.62 $
         \end{tabular} & 
         \begin{tabular}{c}
              $ 76.68 \pm 5.44 $ \\ $ 133.22 \pm 15.95 $ 
         \end{tabular} & 
         \begin{tabular}{c}
              $ 64.22 \pm 1.48 $ \\ $ 1245.92 \pm 11.45 $ 
         \end{tabular} & 
         \begin{tabular}{c}
              \textbf{---}  \\ \textbf{---}  
         \end{tabular} & 
         \begin{tabular}{c}
             $78.46 \pm 14.08$	\\	$\mathbf{8.08 \pm 15.21}$
         \end{tabular} \\\hline
    \end{tabular}
\end{table*}

For CART, the complexity metric is defined as total number of nodes, for SVM, it is defined as total number of feature vectors, for GAM, it is defined as the effective degrees of freedom (EoDF); for GP, it is defined as the total number of internal nodes; and for NLDT, it is defined as total number of variables present in the entire tree.
It is clear that a method with high testing accuracy and low complexity is better. 

The table clearly indicates that NLDT performs well in terms of both metrics. Also, the performance of NLDT scales well with an increase in feature size. CART produces a good compromise on accuracy and complexity, but performs worse than NLDT on both metrics. 
While SVM achieves a high accuracy, in general, the complexity of its classifiers is large, thereby making them not easy to interpret for any explainability purposes. The performance of GP is poor for achieving a high accuracy. GAM is clearly not suitable for problems with a large number of features and cannot be run due to impractical computational time requirement for some problems (marked with a dash). GP cannot match both accuracy and complexity obtained by NLDT. In most problems, NLDT classifiers require fewer conditional rules (albeit with restricted nonlinearities) and still achieve near 100\% correct testing accuracy.

\vspace{-3mm}

\section{Conclusions}\label{sec:conclusions}
In this paper, we have presented three popular binary classification methods -- CART, SVM and GAM. We have also included a genetic programming approach and a recently proposed nonlinear decision tree (NLDT) approach for a comparison with three existing methods on 19 different problems involving two to 500 features. The advantages and disadvantages of each method are described by highlighting one or more problem parameters which control the potential trade-off between complexity of the obtained classifier and its testing accuracy. The extensive comparative results have indicated that the NLDT approach makes an excellent compromise between the testing accuracy and complexity of the classifier. While the former is always important for a classifier, the latter allows an user to look for an explanation involving features and their interactions for the classifier's working principles. 

The study raises a number of interesting future studies: (i) extension to multi-class classification problems, (ii) extension to regression problems, (iii) use of the bilevel approach similar to that used in NLDT search with GP to improve GP's performance, (iv) extension to nonlinear forest (NLF) search involving multiple NLDTs for generating more compact, but slightly more complex and more accurate rules. 

\vspace{-1mm}

\bibliographystyle{plain}
\bibliography{main.bib}
\end{document}